\documentclass[letterpaper, 10 pt, conference]{ieeeconf}
\IEEEoverridecommandlockouts
\usepackage{spconf,amsmath,graphicx}

\usepackage{amssymb}
\usepackage{algorithm}
\usepackage[noend]{algpseudocode}
\usepackage{subfigure}
\usepackage{tabularx}

\usepackage{url}

\title{Crowd Sourcing Image Segmentation with \MakeLowercase{ia}STAPLE}
\name{Dmitrij Schlesinger$^{\star \dagger}$ \ \  Florian Jug$^{\star \ddagger}$ \ \  Gene Myers$^{\ddagger}$ \ \  Carsten Rother$^{\dagger}$ \ \  Dagmar Kainm\"uller$^{\ddagger}$\thanks{
We are grateful to the Pallas Ludens GmbH and especially to D.~Kondermann for providing the  crowd sourcing platform.
Financial support by the BMBF for the competence center for Big Data ScaDS and the 031A099 project is gratefully acknowledged. 
We also thank the European Research Council (ERC, grant agreement No 647769) for their support. Computations were performed at the ZIH at TU Dresden.}
}
\address{
$^{\dagger}$ Computer Vision Lab Dresden, Dresden University of Technology\\
$^{\ddagger}$ Max Planck Institute of Molecular Cell Biology and Genetics\\
$^{\star}$ authors contributed equally
}

\begin{document}
%\ninept
%
\maketitle
\begin{abstract}
We propose a novel label fusion technique as well as a crowdsourcing protocol to efficiently obtain accurate epi\-thelial cell segmentations from non-expert crowd workers. Our label fusion technique simultaneously estimates the true segmentation, the performance levels of individual crowd workers, and an image segmentation model in the form of a pairwise Markov random field. We term our approach image-aware STAPLE (iaSTAPLE) since our image segmentation model seamlessly integrates into the well-known and widely used STAPLE approach. In an evaluation on a light microscopy dataset containing more than $5000$ membrane labeled epithelial cells of a fly wing, we show that iaSTAPLE outperforms STAPLE in terms of segmentation accuracy as well as in terms of the accuracy of estimated crowd worker performance levels, and is able to correctly segment 99\% of all cells when compared to expert segmentations. These results show that iaSTAPLE is a highly useful tool for crowd sourcing image segmentation.
\end{abstract}
\section{Introduction}
Cell segmentation is an important and ubiquitous step in the process of scientific discovery in biology and in clinical applications. 
Ground truth (GT) segmentations of cells in microscopic images are not only needed for biomedical analyses, but also for training and evaluating automated segmentation methods and anatomy models. 
Due to rapidly evolving microscopes, fluorescent dyes, and markers, there is no such thing as a ``default'' appearance of cells in microscopic images. It is therefore common practice to manually generate GT segmentations for each new task to be solved. Since this is a tedious and time consuming task, the amount and quality of available GT is often rather low. 

\enlargethispage{2\baselineskip}
\begin{figure}[htb]
\centering
\includegraphics[width=\linewidth]{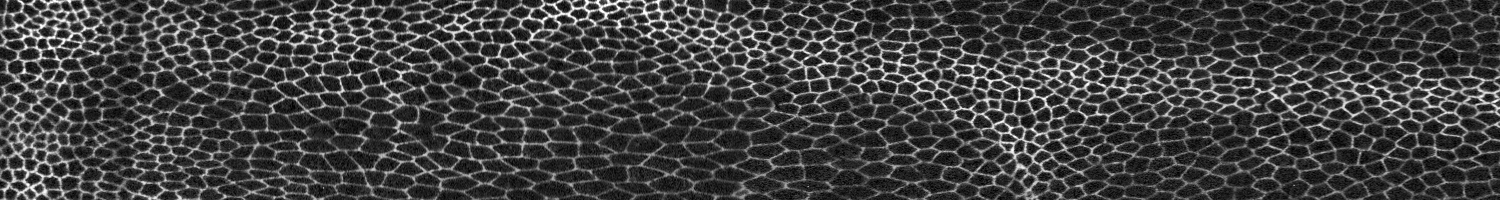}
\vspace{-6mm}
\caption{\label{fig:flywing}Microscopic image of a developing fly wing (crop).}
\end{figure}

% 2) Idea (STAPLE here)
We propose an alternative solution for GT generation: Instead of asking a few (expensive) experts to generate GT, we aim at crowd-sourcing this task to many non-expert workers. Our technique extends the popular STAPLE approach~\cite{warfield2004simultaneous} by simultaneously estimating not only the true segmentation and user performance levels, but jointly also learning an image segmentation model in the form of a pairwise Markovian random field (MRF, cf.\ Section~\ref{sec:method}). We call our new method ``iaSTAPLE'', for \textbf{i}mage-\textbf{a}ware STAPLE. Our method yields ``more informed'' estimates than STAPLE by means of the additional image model, and it enables the segmentation of image regions that were not processed by crowd workers at all. This leads to significantly improved overall performance, as we demonstrate in a quantitative evaluation.

\enlargethispage{2\baselineskip}

%We are not the first to extend STAPLE. 
Landman et al.~\cite{landman2012robust} showed that STAPLE offers a straightforward way to deal with partial as well as multiple segmentations by the same user. Their approach, termed STAPLER, thus reveals an important feature of STAPLE not shown in~\cite{warfield2004simultaneous}. 
In the remainder of this document we will, for simplicity, use the abbreviation STAPLE to refer to both, the original STAPLE~\cite{warfield2004simultaneous} and STAPLER~\cite{landman2012robust}.  
%JUG: Since this was a critical point during MICCAI reviews we said (back then) to put this into the main text of the document. I think we need to prevent readers from not seeing this remark!!!
%\footnote{In further, we use the term STAPLE to refer both the original STAPLE \cite{warfield2004simultaneous} and STAPLER \cite{landman2012robust}, since they are very similar.}. 

The LOP STAPLE approach~\cite{DBLP:journals/tmi/AslHLW14} employs an MRF-based segmentation model. However, as opposed to our proposed method, LOP STAPLE uses a hand-crafted, un-trained MRF model and relies on approximate mean field based inference. 
Non-local STAPLE~\cite{DBLP:conf/miccai/AsmanL12} shares our idea of making use of image intensities. However, intensities are not exploited in the form of an appearance model for segmentation as we propose, but in a different context, namely to improve the registration of atlases. 
% DIMA: not sure what the next sentence means -- if not necessary I would leave it out.
%Beside of that, an over-simplified independent segmentation model is used here. 
The approach closest to our own work is probably iSTAPLE~\cite{DBLP:conf/miip/LiuMTS13}, where image intensities are used as an appearance model for segmentation. However, iSTAPLE employs a pixel-independent segmentation model as well as a simple Gaussian probability distribution of gray-values as appearance model. Our method, in contrast, uses a learned pairwise MRF for segmentations and a Gaussian Mixture model with shading to model appearances.

% I omitted "Local MAP STAPLE (Commowick et al., 2012) -- Estimating A Reference Standard Segmentation with Spatially Varying Performance Parameters: Local MAP STAPLE Olivier Commowick∗ †, Alireza Akhondi-Asl†, Simon K. Garfield†", because I think it is very far from ours, hence not very relevant -- just for space.

Our main contributions are: $(i)$~A novel crowd-sourcing workflow for efficiently segmenting all cells in membrane labeled epithelia (cf.\ Section~\ref{sec:crowd}). $(ii)$~A generic label fusion technique that can be used in the context of many image segmentation problems. We demonstrate that this is a viable approach for the exemplary application of segmenting membrane labeled epithelial cells in fluorescence microscopy images (cf.\ Figure~\ref{fig:flywing}). Furthermore, we show quantitatively that iaSTAPLE's user performance estimates are more accurate than those of STAPLE/STAPLER. This finding suggests that our user performance estimates may serve as valuable input for automated quality control in crowd-sourced image segmentation, complementing supplementary information like click behavior and performance on test tasks (see e.g.~\cite{sameki2015predicting}). 

%%%%%%%%%%%%%%%%%%%%%%%%%%%%%%%%%%%%%%%%%%%%%%%%%%%%%%%%%%%%%%%%%%%%%%%%%%%%%%%%%%%%%%%%%%%%%%%%%%%%%%%%%%%%%%%

\section{Crowdsourcing Epithelial Cell Segmentation}
\label{sec:crowd}
\enlargethispage{2\baselineskip}

Microscopy images of cell epithelia often contain thousands of cells. Epithelial cells are densely packed with no space in-between. Hence, cell membranes appear as a honeycomb-like polygonal mesh that covers the whole image. Cell size varies considerably, with a scale factor of $10$ not being unusual. Figure~\ref{fig:flywing} shows part of a fly wing. The cell membranes appear bright due to a fluorescent membrane marker that is genetically introduced into the organism. 

\begin{algorithm}[htb]
\caption{\label{alg:crowd}Crowdsourcing Epithelial Cell Segmentation.}
{\fontsize{8}{9}\selectfont
\begin{algorithmic}

\State {\bf Input:} Set $T$ of image tiles $t$, outer loop iterations $K$.

\State {\bf Output:} $P$, a set of closed polygons $p$ describing segmented cell outlines.

\vspace{1ex}

\State $P=\emptyset$;

\For{$k=1\ldots K$}

	\State $T_k=T$ \Comment \begin{minipage}{0.54\linewidth}The set of image tiles not fully covered at the current iteration of the outer loop.\end{minipage}
	\For{$t\in T_k$} $P_t=\emptyset$ \Comment The current set of polygons for each tile. \EndFor
	
	\Repeat
		\For{ $t\in T_k$ }
			$P_t=\Call{CrowdAnnotate}{t,P_t}$
		\EndFor
		\State $T_k=\Call{GetNotCoveredTiles}{T,\bigcup_{t\in T} P_t}$
	\Until{ $T_k=\emptyset$ \textbf{or} $T_k$ did not change.}
	\State $P=P\cup\bigcup_{t\in T} P_t$
	 
\EndFor
	 
\vspace{1ex}

\Function{CrowdAnnotate}{$t, P_t$}
	\If { more than 20 visible cells in $t$ are not yet in $P_t$}
    		\State Outline 20 fully visible cells that are not yet in $P_t$.
   	\Else \ Outline all fully visible cells that are not yet in $P_t$.
	\EndIf
	\State \Return $P_t\cup $ newly drawn polygons. 
\EndFunction

\vspace{1ex}
	 
\Function{GetNotCoveredTiles}{$T,P$}
	\State Mark areas covered by polygons $P$ as foreground in binary image $c$.
	\State Dilate foreground of $c$ by known membrane width. 
	\State $T_{nc}=\emptyset$ \Comment The set of not fully covered tiles to be returned.
	\For{$t\in T$}
	\If{there are more background pixels in $t$ than a threshold} 
		\State $T_{nc}=T_{nc}\cup t$
	\EndIf
	\EndFor
    \State \Return $T_{nc} $

\EndFunction

\end{algorithmic}
}
\end{algorithm}

An individual epithelial cell can be segmented easily by drawing its outline as a closed polygon. We use Amazon Mechanical Turk as crowd sourcing platform, where workers are paid by completed task and not by the hour. Hence all tasks given to the crowd should, by design, be similarly time consuming. To satisfy this criterion in the face of highly varying cell sizes, we define a task to be to outline a fixed number of cells in a given image tile of comfortably visualizable size. 
The respective instruction as given to crowd workers can be found at 
\mbox{\url{bioimagecomputing.com/crowd-instructions}}.

In Algorithm~\ref{alg:crowd} we describe a workflow for yielding (almost) complete coverage of the whole image by means of these tasks performed on overlapping image tiles of fixed size. The inner \textbf{repeat}-loop aims at covering the whole image once, while the outer loop can be used to obtain multiple coverages. 
%
%Given the resulting set of polygonal cell outlines, $P$, we convert each such polygon into a binary segmentation of a small image region: Pixels covered by the outline are marked as \emph{cell membrane}, where the outline is assumed to have the known membrane width. The polygon interior is marked as \emph{cell interior}. Resulting segmentations are partial, as they do not cover any pixels that are outside of the respective polygon. We apply our novel label fusion method, iaSTAPLE (cf.\ Sec.\ \ref{sec:method}), to the case of binary foreground (membrane) vs. background (cell interior) segmentations given such a set of crowd annotations. 

%%%%%%%%%%%%%%%%%%%%%%%%%%%%%%%%%%%%%%%%%%%%%%%%%%%%%%%%%%%%%%%%%%%%%%%%%%%%%%%%%%%%%%%%%%%%%%%%%%%%%%%%%%%%%%%

\enlargethispage{2\baselineskip}
\section{\MakeLowercase{ia}STAPLE}
\label{sec:method}

{\bf Model}. Let $x$ be an observation (image) and $y$ be a hidden variable, i.e.~the binary segmentation in our case. Formally, the labeling is a mapping $y:R\rightarrow L$, $L=\{0,1\}$ (called label set), that assigns $0$ (background) or $1$ (foreground) to each pixel $i\in R$. By $y_i\in L$, we will denote the label chosen in the pixel $i$. Analogously, the image $x:R\rightarrow C$ is a mapping assigning a gray value $c\in C$ to each pixel. The gray value of pixel $i$ is denoted by $x_i$. Let us denote user inputs by $z^u:R\rightarrow L$, where $u$ indicates a particular user, so if we have $m$ users, $u=1\ldots m$. Our key idea is to consider user inputs as ``additional observations'' -- the $u$-th user ``observes'' the true scene $y$ and gives his/her opinion about the label $l\in L$ of each pixel $i\in R$. It is also reasonable to assume that users are conditionally independent from each other given a labeling $y$. To summarize, the joint probability distribution for all model constituents is\footnote{Parameters are separated from random variables by a semicolon.}
\begin{eqnarray}\label{eq:model}
\lefteqn{p(x,y,z^1,\ldots z^m;\theta_p,\theta_a,p_u,u{=}1{\ldots}m)=}\nonumber \\
& & =p(y;\theta_p)\cdot p(x|y;\theta_a)\cdot\prod_{u=1}^m p_u(z^u|y) ,
\end{eqnarray}
with $\theta_p$ the parameters of the prior probability distribution of labelings $p(y;\theta_p)$, and $\theta_a$ the parameter of the gray value model $p(x|y;\theta_a)$. Each user is characterized by his own, initially unknown probability distribution $p_u(z^u|y)$ reflecting his/her \emph{reliability}.
Let us consider all model parts in more detail.

Our prior model $p(y;\theta_p)$ is a Markovian Random Field (MRF) over the graph $G=(R,E)$, where the node set $R$ corresponds do the pixel grid and the edge set $E$ reflecting pixel neighborhoods. The associated energy is
\begin{equation}\label{eq:en1}
E(y,\theta_p)=\sum_{i\in R} \psi_0(y_i)+\sum_c\sum_{ij\in E_c} \psi_c(y_i,y_j) .
\end{equation}
Unary potentials $\psi_0:L\rightarrow\mathbb R$ assign values to each label $l\in L$ of each pixel $i\in R$. Pairwise potentials are defined in a densely connected neighborhood, where edges $E$ are partitioned into \emph{edge classes} $E_c$. Classes are characterized by the translational vector that connects the corresponding pixels (see e.g.\  \cite{FlachS11,nowozin2011decision}). Pairwise potentials $\psi_c:L\times L\rightarrow \mathbb R$ are class specific and shared by all edges of the same class. Hence the free model parameters $\theta_p$ are the unary potentials $\psi_0$, and the pairwise potentials $\psi_c$ for all $c$, and the prior probability distribution is
\begin{eqnarray}\label{eq:p1}
\lefteqn{p(y;\theta_p)=\frac{1}{Z(\theta_p)}\exp\bigl[-E(y,\theta_p)\bigr] ,}\nonumber\\
& & \text{with} \ \ Z(\theta_p)=\sum_y \exp\bigl[-E(y,\theta_p)\bigr] .
\end{eqnarray}
Like STAPLE~\cite{warfield2004simultaneous}, we assume conditional independence of user inputs at each pixel, given a labeling $y$. So $p_u(z^u|y)=\prod_i p_u(z^u_i|y_i)$, with $z^u_i$ denoting the label given by user $u$ in the pixel $i$. Hence, the unknown parameters of user models are user specific conditional probability distributions $p_u:L\times L\rightarrow\mathbb R$, with $p_u(l'|l)\geq 0$ and $\sum_{l'}p_u(l'|l)=1$ for all $l\in L$. The key difference between our approach and STAPLE is that our model is aware of the input image. At each pixel we use a Gaussian Mixture to model the gray value distribution, with label specific weights. In addition, we use a so-called shading field~\cite{Schles2008} that allows smoothly varying deviations of the appearance model. Formally, the shading is a mapping $s:R\rightarrow\mathbb C$, like the image itself, yet it is required to be spatially smooth. Given a shading field $s$ and a labeling $y$, the appearance model is
\begin{equation}\label{eq:xmodel}
p(x|y,s)=\prod_i\sum_j w_{y_ij}\frac{1}{\sqrt{2\pi}\sigma}\exp\left[-\frac{\bigl((x_i-s_i)-\mu_j\bigr)^2}{2\sigma^2}\right] ,
\end{equation}
with the prior probability distribution for shading being a Gaussian MRF, $\mu_j$ being the centers of Gaussians, a common $\sigma$ to all Gaussians, $w_{l,j}$ being label specific weights, and the initially unknown shading $s$.

\vspace{1ex}\noindent
\enlargethispage{2\baselineskip}
{\bf Learning.} We formulate the learning problem according to the Maximum Likelihood principle and describe here the learning of user models $p_u$. All other model components are learned as in~\cite{FlachS11,Schles2008}. Given an image $x$ and user annotations $z^u$, $u=1\ldots m$, the task is to maximize the logarithm of their joint probability, which is obtained by marginalization over hidden $y$:
\begin{eqnarray}
\lefteqn{\ln p(x,z^1,\ldots,z^m;\theta_p,\theta_a,p_u,u{=}1{\ldots}m)=} \nonumber \\
& & = \ln \sum_y \left[\frac{1}{Z(\theta_p)} \exp\bigl[-E(x,y,\theta_p,\theta_a)\bigr]\cdot\right.\\
& & \left.\cdot\prod_{u}\prod_i p_u(z^u_i|y_i)\right]\rightarrow\max_{p_u,u{=}1{\ldots}m} ,
\end{eqnarray}
where $E(x,y,\theta_p,\theta_a)$ summarizes all energy terms that are fixed for the moment, i.e.~the prior energy \eqref{eq:en1} and logarithms of \eqref{eq:xmodel}. We use Expectation Maximization as follows: $(i)$ In the E-step the marginal posterior label probabilities 
\begin{equation}\label{eq:marg}
p(y_i{=}l|x,z_1,\ldots,z_m)=\sum_{y:y_i=l} p(y|x,z_1,\ldots,z_m)
\end{equation}
for each pixel and each label are computed, followed by $(ii)$, the M-step, where the following optimization problem is solved:
\begin{equation}\label{eq:userm}
\sum_i\sum_l\sum_u p(y_i{=}l|x,z^1,\ldots,z^m)\cdot \ln p_u(z^u_i|l)\rightarrow\max_{p_u,u{=}1{\ldots}m} .
\end{equation}
Let $I_{ul'}$ be the set of pixels that user $u$ assigned label $l'$. Probabilities $p_u(l'|l)$ solving \eqref{eq:userm} are proportional to the \emph{frequencies} $p_u(l'|l)\propto n_u(l',l)$, where
\begin{equation}\label{eq:freq}
n_u(l',l)=\sum_{i\in I_{ul'}} p(y_i{=}l|x,z^1,\ldots,z^m) .
\end{equation}
Since the computation of marginal probabilities in \eqref{eq:marg} is intractable for general MRFs, we use Gibbs Sampling: We sample labelings $\hat y$ according to the posterior $p(y|x,z^1,\ldots,z^m)$, and count how many times which label was generated at which pixel. Doing so, we observe the following: Usually, it is necessary to have many independent samples $\hat y$ drawn from the target probability distribution $p(y|x,z^1,\ldots,z^m)$ in order to estimate the frequencies of \eqref{eq:freq}. Our frequencies, however, are sufficient statistics of low order, i.e.~for a particular user we need to estimate just two values $p_u(0|1)$ and $p_u(1|0)$. Hence, in order to accelerate the overall learning procedure we use \emph{warm starts}, meaning that we use the sample obtained at the previous iteration of the EM-algorithm as the initialization and perform just one Gibbs Sampling iteration. Similar learning procedures are for example known as Persistent Contrastive Divergence~\cite{tieleman2008training}. The learning algorithm is summarized in Algorithm~\ref{alg:alg1}.

\begin{algorithm}[tbh]
\caption{\label{alg:alg1}Learning the User Models.}
{\fontsize{8}{9}\selectfont
\begin{algorithmic}

\State {\bf Input:} 
\begin{minipage}[t]{0.9\linewidth}
Image $x$, user inputs $z^u$, prior model parameters $\theta_p=(\psi_0,\psi_c)$, appearance model $\theta_a=(w,\mu,\sigma)$, initial user models $p_u^{(0)}$, initial random labeling $\hat y^{(0)}$.
\end{minipage}

\State {\bf Output:} 
\begin{minipage}[t]{0.8\textwidth}
Learned user models $p_u$.
\end{minipage}

\vspace{1ex}
\For{ $t := 0\ldots T$ }

	\State \begin{minipage}{0.9\linewidth} Perform one iteration of Gibbs Sampling to draw a labeling from the posterior 
	probability distribution $p(y|x,z^1,\ldots,z^m)$, starting from the labeling $\hat y^{(t)}$, using the 
	current model parameters $(\theta_p,\theta_a,p_u^{(t)},u=1\ldots m)$ $\rightarrow$ obtain the new labeling $\hat y^{(t+1)}$.
	\end{minipage}
	
	\vspace{1ex}
	\For{ all users $u$ }
		\State Compute frequencies $n_u(l',l)=\#(z^u_i{=}l',\hat y^{(t+1)}_i{=}l)$, $\forall l',l$.

		\State Perform the step toward the optimal value with a step size $\lambda$:
		
		\vspace{-3ex}
		\begin{equation*}
		p_u^{(t+1)}(l'|l)=(1-\lambda)\cdot p_u^{(t)}(l'|l)+\lambda\cdot \frac{n_u(l',l)}{\sum_{l''}n_u(l'',l)}, \ \ \forall l',l
		\end{equation*}

		\vspace{-2ex}
	\EndFor
\EndFor
\end{algorithmic}
}
\end{algorithm}

\vspace{1ex}\noindent
{\bf Inference}. 
We compute the Maximum Posterior Marginal decision (MPM) by computing the node-wise posterior marginal probabilities, followed by choosing the labelling which has the best marginal at each node. We do this, again, by using Gibbs Sampling. 

%%%%%%%%%%%%%%%%%%%%%%%%%%%%%%%%%%%%%%%%%%%%%%%%%%%%%%%%%%%%%%%%%%%%%%%%%%%%%%%%%%%%%%%%%%%%%%%%%%%%%%%%%%%%%%%

\section{Results and Discussion}
\enlargethispage{2\baselineskip}

\begin{figure}[htb]
\centering
\subfigure[Pixel accuracy]{\includegraphics[width=0.32\linewidth]{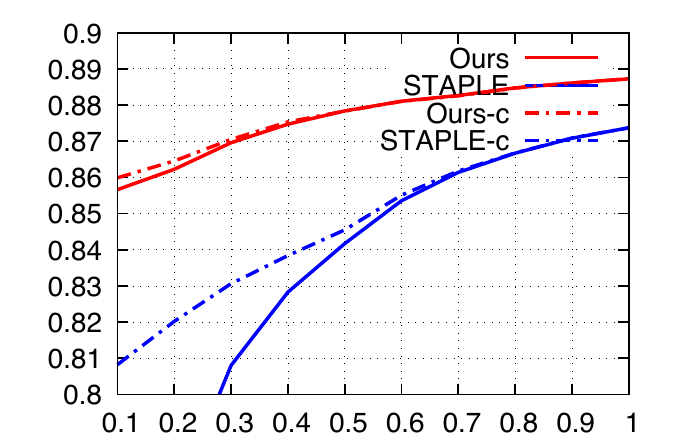}}
\subfigure[F1-score]{\includegraphics[width=0.32\linewidth]{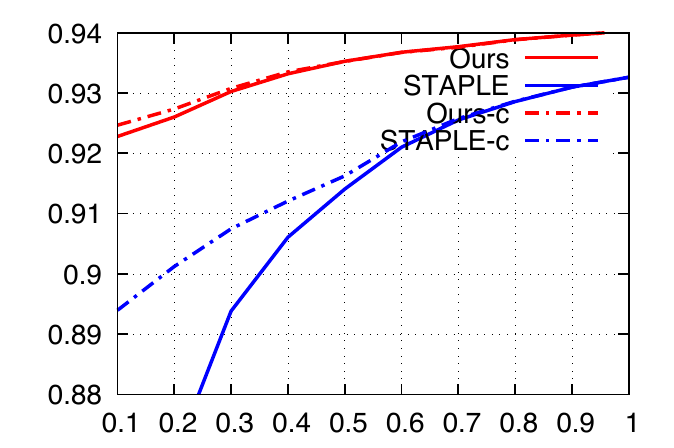}}
\subfigure[VoI ($\times 10^{-6}$)]{\includegraphics[width=0.32\linewidth]{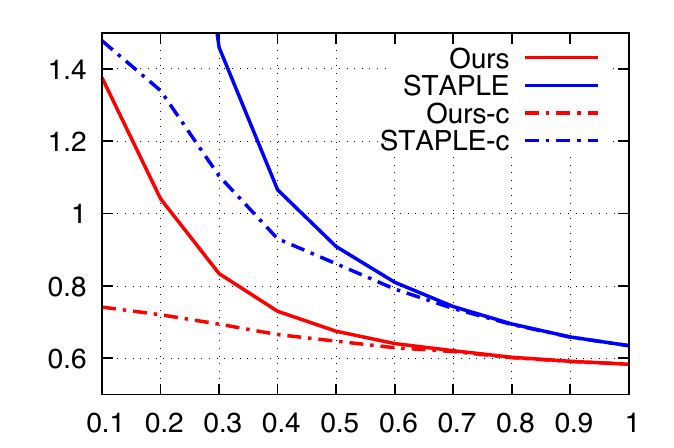}}
\subfigure[under-segmentation]{\includegraphics[width=0.49\linewidth]{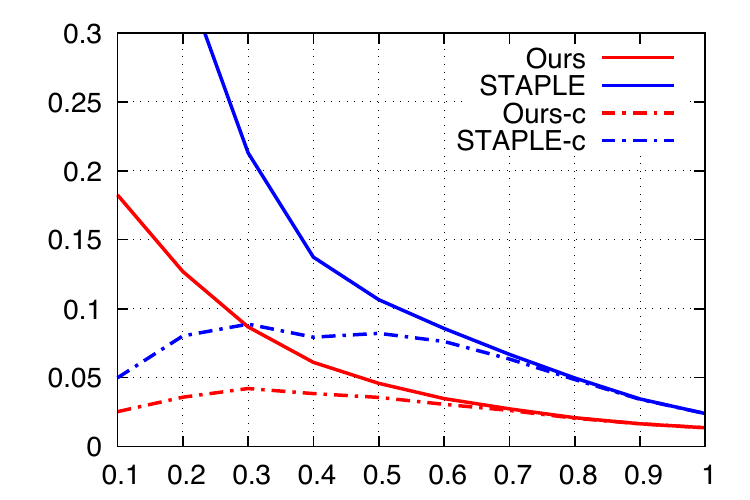}}
\subfigure[over-segmentation]{\includegraphics[width=0.49\linewidth]{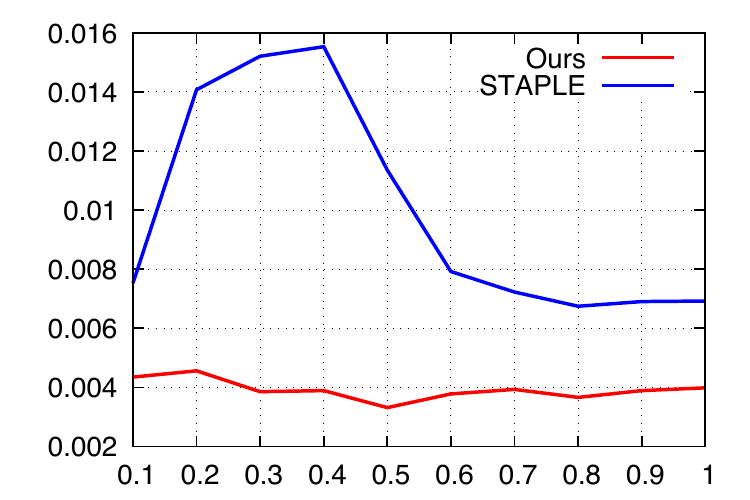}}
\ ~\vspace{-4mm}
\caption{\label{fig:res}Comparison of the estimated segmentations. The suffix ``-c'' indicates that we evaluated the quality only on regions that are covered by at least one user annotation.}
\end{figure}

{\bf Segmentation Evaluation}. 
To evaluate iaSTAPLE, we randomly pick certain fractions of crowd annotations, ranging from 10\% to 100\% of $1584$ annotations obtained for a $5000$-cell image of a developing fly wing. Given such a fraction, we train our model and estimate the true segmentation. We compare results to a manually prepared ground truth (GT)
%\footnote{The ground truth was manually produced by an expert.}
according to three standard error measures, namely $(i)$~pixel accuracy, $(ii)$~F1-score, and $(iii)$~variation of information (VoI). To compute the VoI, we partition the image by assessing connected components of the ``interior'' label and assigning all ``membrane'' pixels to the closest connected component. We repeat each experiment $10$ times per given fraction and report respective average errors. We perform the same experiments for STAPLE/STAPLER. Fig.~\ref{fig:res} (top row) shows resulting errors as a function of the respective fraction of annotations. Since STAPLE/STAPLER cannot handle regions that are not annotated by at least one user, we also evaluate all results exclusively on image areas that are covered by at least one annotation (see ``STAPLE-c'' and ``Ours-c'' in Fig.~\ref{fig:res}). 
%\begin{figure}[htb]
%\centering
%\subfigure[Pixel accuracy]{\includegraphics[width=0.32\linewidth]{./images/accuracy.pdf}}
%\subfigure[F1-score]{\includegraphics[width=0.32\linewidth]{./images/f1.pdf}}
%\subfigure[VoI ($\times 10^{-6}$)]{\includegraphics[width=0.32\linewidth]{./images/varinf.pdf}}
%\vspace{-3mm}
%\caption{\label{fig:res}Comparison of the estimated segmentation. The suffix ``-c'' indicates that we evaluated the quality only on regions that are covered by at least one user annotation.}
%\end{figure}
%
%\begin{figure}[htb]
%\subfigure[under-segmentation]{\includegraphics[width=0.49\linewidth]{./images/under.pdf}}
%\subfigure[over-segmentation]{\includegraphics[width=0.49\linewidth]{./images/over.pdf}}
%\caption{\label{fig:resunder}
%Percentages of under- and over-segmented cells. The suffix ``-c'', again, indicates that only image regions with given user annotations were analyzed.}
%\end{figure}

To complement the standard error measures by easily interpretable yet non-standard measures, we also assess \emph{under-} and \emph{over-segmentations}. An under-segmentation occurs if two neighboring GT cells are merged by mistake, an over-segmentation if a GT cell is split into two by mistake. We count under- and over-segmentations as follows: We first compute all connected background components (cells) from results and from GT. Then we perform a Hungarian Matching between the two sets of cells, where cells are allowed to be matched only if they are reasonably close. Unmatched cells in the segmentation results count as over-segmentations, while unmatched cells in the GT count as under-segmentations. In Figure.~\ref{fig:res} (bottom row) we show percentages of under- and over-segmented GT cells as a function of the fraction of crowd annotations used in the experiment. 

iaSTAPLE outperforms STAPLE/STAPLER in all experiments and all error measures, for partial image coverage as well as for full coverage. For most error measures, iaSTAPLE's segmentation accuracy surpasses STAPLE/STAPLER already when using less than $50$\% of the respective amount of crowd user inputs. In case of 100\% crowd annotations, the total percentage of wrongly segmented GT cells for iaSTAPLE is only about 1\%. We hypothesize that this accuracy is sufficient for iaSTAPLE's results to serve as ground truth for many practical applications. However, future work has to assess the respective inter-expert variability, or alternatively the impact of expert-GT vs crowd-GT on application-specific measures, like e.g.\ biological quantities of interest, or performance of GT-trained segmentation algorithms (cf.\ \cite{maier2014can}). 
\begin{figure}[htb]
\begin{center}
	\includegraphics[width=0.32\linewidth]{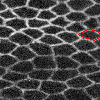}
	\includegraphics[width=0.32\linewidth]{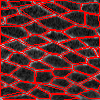}
	\includegraphics[width=0.32\linewidth]{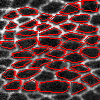}	
	
	\vspace{0.5ex}
	\includegraphics[width=0.32\linewidth]{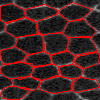}
	\includegraphics[width=0.32\linewidth]{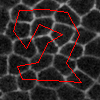}
	\includegraphics[width=0.32\linewidth]{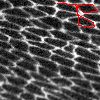}
\end{center}
\ ~\vspace{-11mm}
\caption{\label{fig:resu} Exemplary user annotations (red polygons overlaid with original images). Bottom row, left to right: Overall best, a very bad, and overall worst annotation.}
\end{figure}

\sloppy

\vspace{1ex}\noindent
{\bf User Rating}.
\enlargethispage{2\baselineskip} 
iaSTAPLE, like STAPLE/STAPLER, estimates performance measures for each user and label, $p_u(z^u_i{=}\text{membrane}|y_i{=}\text{cell})$ and $p_u(z^u_i{=}\text{cell}|y_i{=}\text{membrane})$. To capture the quality of users with a single number we propose to use the user specific pixel accuracy, i.e.~the number of correctly labeled pixels divided by the number of all pixels a user has annotated. To give an impression of the effectiveness of this measure, Fig.~\ref{fig:resu} shows the annotations by the respective best user (bottom left) as well as worst user (bottom right). 

We evaluate estimated user ratings as follows: Let $r_u$ be the rank of user $u$ among all users in terms of the fraction of pixels where $u$ agrees with the segmentation estimate yielded by STAPLE or iaSTAPLE. Let $r_u^\ast$ be the rank of user $u$ in terms of the fraction of pixels where $u$ agrees with the ground truth segmentation. Thus, $r_u$ represents an estimated user ranking yielded by STAPLE or iaSTAPLE, while $r_u^\ast$ represents the respective true user ranking. The average absolute difference between estimated and true user ranking can be interpreted as the \emph{quality} of the respective estimated user ranking, where lower (i.e.\ more similar to true ranking) is better; formally, $1/m \sum_{i=1}^m |r_u-r_u^\ast|$. Considering the whole set of user annotations ($179$ users), iaSTAPLE achieves a quality value of $8.63$, while STAPLE achieves a value of $10.06$. Hence in these terms iaSTAPLE estimates a more accurate user ranking than STAPLE. This result suggests that iaSTAPLE's user rating estimates are more useful than STAPLE's for improving automated quality control in crowd-sourced image segmentation. 

%%%%%%%%%%%%%%%%%%%%%%%%%%%%%%%%%%%%%%%%%%%%%%%%%%%%%%%%%%%%%%%%%%%%%%%%%%%%%%%%%%%%%%%%%%%%%%%%%%%%%%%%%%%%%%%
\enlargethispage{2\baselineskip}
\section{Conclusion}
We proposed iaSTAPLE, a novel image-aware method for label fusion, as well as an efficient crowd-sourcing workflow for epithelial cell segmentation. We applied this workflow on microscopy images of membrane labeled fly wing epithelia, and compared the results of iaSTAPLE against the well-known STAPLE. We showed that iaSTAPLE outperforms STAPLE/STAPLER both in terms of accuracy of the estimated segmentation, as well as in the quality of the estimated user ranking. 
These results make iaSTAPLE the method of our choice for crowd-sourced image segmentation.

\bibliographystyle{IEEEbib}
\bibliography{isbi2017,AllPapersJug}

\end{document}